\documentclass[10pt]{llncs}
\usepackage{rotating}
\usepackage{pdflscape}
\usepackage{amsmath}
\usepackage{amsfonts}
\usepackage{amssymb}
\usepackage{amstext}
\usepackage{latexsym,stmaryrd}

\usepackage{inferance}
\usepackage{deduc}
\def\infty{\text{BUG}}

\renewcommand{\leq}{\mathrel{\leqslant}}
\renewcommand{\geq}{\mathrel{\geqslant}}

\newcommand{\lconj}{\ensuremath{\mathbin{\wedge}}}
\newcommand{\ldisj}{\ensuremath{\mathbin{\vee}}}

\renewcommand\ruleformat[1]{\ #1}

\begin{document}
\title{A Mining-Based Compression Approach for Constraint Satisfaction Problems} 
%\title{A Mining-Based Method for Size-Reduction of CNF Boolean Formulae}
%\titlerunning{}
\author{Said Jabbour and Lakhdar Sais and Yakoub Salhi}
\institute{CRIL - CNRS, University of Artois, France}
\maketitle

\begin{abstract}
 In this paper, we propose an extension of our Mining for SAT framework to Constraint satisfaction Problem (CSP). We consider n-ary extensional constraints (table constraints).  Our approach aims to reduce the size of the CSP by exploiting the structure of the constraints graph and of its associated microstructure. More precisely, we apply itemset mining techniques to search for closed frequent itemsets on these two representation. Using Tseitin extension, we rewrite the whole CSP to another compressed CSP equivalent with respect to satisfiability. Our approach contrast with previous proposed approach by Katsirelos  and Walsh, as we do not change the structure of the constraints.  

\end{abstract}

\section{Introduction}
\label{sect:introduction}

The table constraint is considered for a long time as particularly important in constraint satisfaction problems (CSP). Indeed, on of the most used formulation of CSP consists in expressing the each constraint in extension or as a relation among variables with associated finite domains. Many research work, consider table constraints as the standard representation. Indeed, any constraint can be expressed using a set of allowed or forbidden tuples. However, the size of these kind of extensional constraints might be exponential in the worst case. In \cite{KatsirelosW07},  Katsirelos and Walsh proposed for the first time a compression algorithm for large arity extensional constraints. The proposed algorithm attempts to capture the structure that may exist in a table constraint. The authors proposed an alternative representation of the set of tuples of a given relation by a set of compressed tuples. The proposed representation may lead to an exponential reduction in space complexity. However, the compressed tuples may be larger than the arity of the original constraint. Consequently, the obtained CSP do not follow the standard representation of the table  constraint. The authors use decision trees to derive a set of compressed tuples.  

In this paper, we present a new compression algorithm that combines both  itemset mining techniques and Tseitin extension principles to derive a new compact representation of the table constraints. First, we show our previous Mining for SAT approach can be extended to deal with the CSP by considering the constraint graph as a transaction database, where the transactions corresponds to the constraints and items to the variables of the CSP. The closed frequent itemsets corresponds to subset of variables shared most often by the different constraint of the CSP. Secondly, using extension (auxiliary variables) we show how such constraints can be rewritten while preserving satisfiability. Secondly, we consider each table constraint individually, we derive a new transaction database made of a sequence of tuples i.e. a set of indexed tuples. More precisely, each value of a tuple is indexed with its position in the constraint. By enumerating closed frequent itemsets on such transaction database, we are able to search for the largest rectangle in the table constraint. Similarly, with extension principle, we show how such constraint can be compressed while preserving the traditional representation. 

\section{Technical background and preliminary definitions}
\subsection{Frequent Itemset Mining Problem}
\label{sec:im}
%\subsection{Preliminary Notations and Definitions}

Let $\cal I$ be a set of {\it items}. A set $I\subseteq {\cal I}$ is called an {\it itemset}.
A {\it transaction} is a couple $(tid,I)$ where $tid$ is the {\it transaction identifier} and 
$I$ is an itemset. A {\it transaction database} ${\cal D}$ is a finite set of transactions over $\cal I$
where for all two different transactions, they do not have the same transaction identifier.
We say that a transaction $(tid, I)$ {\it supports} an itemset $J$ if  $J\subseteq I$.

The {\it cover} of an itemset $I$ in a transaction database $\cal D$ is the set of
identifiers of  transactions in   $\cal D$ supporting $I$: 
${\cal C}(I,{\cal D})=\{tid\mid (tid,J)\in{\cal D}\texttt{ and }I\subseteq J \}$.
The {\it support} of an itemset $I$ in $\cal D$ is defined by:
${\cal S}(I,{\cal D})=\mid {\cal C}(I,{\cal D})\mid$.
Moreover, the {\it frequency} of $I$ in $\cal D$ is defined by:  
${\cal F}(I,{\cal D})=\frac {{\cal S}(I,{\cal D})}{\mid{\cal D}\mid}$.

For example, let us consider  the transaction database in Table~\ref{tty}.
Each transaction corresponds to the favorite writers of a library member.
For instance, we have ${\cal S}(\{Hemingway,  Melville\},{\cal D})=|\{002,004\}|=2$ and 
${\cal F}(\{Hemingway, $\\ $ Melville\},{\cal D})=\frac{1}{3}$.

\begin{table}
\label{tab:td}
\centering{
{\small
\begin{tabular}{|c|c|}
\hline
tid & itemset \\
\hline
001 &  $Joyce,  Beckett, Proust$	\\
\hline
002 & $Faulkner, Hemingway,  Melville$\\	
\hline
003 &  $Joyce, Proust$\\
\hline
004 & $Hemingway,  Melville$ \\
\hline
005 & $Flaubert, Zola$ \\
\hline
006 & $Hemingway, Golding$ \\
\hline
\end{tabular}}
}
\caption{An example of transaction database ${\cal D}$ }
\label{tty}
\end{table}

\noindent Let $\cal D$ be a transaction database over $\cal I$ and $\lambda$ a minimal support threshold.
The frequent itemset mining problem consists of computing the following set:
${\cal FIM}({\cal D},\lambda)=\{I\subseteq {\cal I}\mid {\cal S}(I,{\cal D})\geq\lambda\}$.

The problem of computing the number of  frequent itemsets is $\# P$-hard~\cite{Gunopulos2003}. The complexity class $\# P$ 
corresponds to the set of counting problems associated with a decision problems in $NP$.
For example, counting the number of models satisfying a CNF formula is a $\# P$  problem.

% \begin{proposition}[Anti-monotonicity]
% \label{prop:AM}
% Let $I$ and $I'$ be two itemsets such that $I\subseteq I'$.
% If $I'\in {\cal FIM}({\cal D},\lambda)$ then $I\in {\cal FIM}({\cal D},\lambda)$.
% \end{proposition}
 
%\subsection{Maximal and Closed Frequent Itemsets}
%\label{subsec:maxclo}
Let us now define two condensed  representations of the set of all frequent itemsets: 
maximal and closed frequent itemsets. 

\begin{definition}[Maximal Frequent Itemset]
Let $\cal D$ be a transaction database, $\lambda$ a minimal support threshold
and  $I\in{\cal FIM}({\cal D},\lambda)$. $I$ is called maximal when for all $I'\supset I$, 
$I'\notin {\cal FIM}({\cal D},\lambda)$ ($I'$ is not a frequent itemset). 
\end{definition}

We denote by ${\cal MAX}({\cal D}, \lambda)$ the set of all maximal frequent itemsets
in $\cal D$ with $\lambda$ as a minimal support threshold.
For instance, in the previous example, we have ${\cal MAX}({\cal D}, 2)=\{\{Joyce, Proust\},\{Hemingway,  Melville\}\}$. 
 
\begin{definition}[Closed Frequent Itemset]
Let $\cal D$ be a transaction database, $\lambda$ a minimal support threshold
and  $I\in{\cal FIM}({\cal D},\lambda)$. $I$ is called closed when for all $I'\supset I$, 
${\cal C}(I,{\cal D})\neq {\cal C}(I',{\cal D}) $. 
\end{definition}

We denote by ${\cal CLO}({\cal D}, \lambda)$ the set of all closed frequent itemsets
in $\cal D$ with $\lambda$ as a minimal support threshold.
For instance,  we have ${\cal CLO}({\cal D}, 2)=\{\{Hemingway\},$\\$\{Joyce, Proust\},\{Hemingway,  Melville\}\}$.
In particular, let us note that we have ${\cal C}(\{Hemingway\},{\cal D})=\{002,004,006\}$ and ${\cal C}(\{Hemingway,Melville\},{\cal D})=\{002,004\}$. 
That explains why $\{Hemingway\}$ and $\{Hemingway,  Melville\}$ are both closed.
One can easily see that if all the closed (resp. maximal) frequent itemsets are computed, 
then all the frequent itemsets can be computed without using the corresponding database. 
Indeed, the  frequent itemsets correspond to all the subsets of the closed (resp. maximal) frequent itemsets. \\

Clearly, the number of maximal (resp. closed) frequent itemsets is  significantly smaller
than the number of frequent itemsets. Nonetheless, this number is not always polynomial in the size of the database~\cite{Yang2004}. 
In particular, the problem of counting the number of maximal frequent itemsets 
is $\#P$-complete (see also~\cite{Yang2004}).

Many algorithm has been proposed for enumerating frequent closed itemsets. One can cite Apriori-like algorithm, originally proposed in \cite{agrawal93} for mining frequent itemsets for association rules. It proceeds by a level-wise search of the elements of  ${\cal FIM}({\cal D},\lambda)$.
Indeed, it starts by computing the elements of  ${\cal FIM}({\cal D},\lambda)$ of size one.
Then, assuming the element of ${\cal FIM}({\cal D},\lambda)$ of size $n$ is known,  
it computes a set of candidates  of size $n+1$ so that $I$ is a candidate if and only if 
all its subsets are in  ${\cal FIM}({\cal D},\lambda)$. This procedure is iterated until no more candidates are found. 
Obviously, this basic procedure is enhanced using some properties such as the anti-monotonicity property that allow us to reduce the search space. Indeed, if $I\notin{\cal FIM}({\cal D},\lambda)$, then $I'\notin{\cal FIM}({\cal D},\lambda)$ for all  $I'\supseteq I$. In our experiments, we consider one of the state-of-the-art algorithm LCM for mining frequent closed itemsets proposed by Takeaki Uno et al. in \cite{UnoKA04}. In theory, the authors prove that LCM exactly enumerates the set of frequent closed itemsets within polynomial time per closed itemset in the total input size.  Let us mention that LCM algorithm obtained the best implementation award of FIMI'2004 (Frequent Itemset Mining Implementations). 

\subsection{Constraint Satisfaction Problems: Preliminary definitions and notations}
A constraint network  is defined as a tuple $\mathcal{P} =
\langle\mathcal{X}, \mathcal{D}, \mathcal{C}\rangle$. $\mathcal{X}$ is a finite set of $n$ variables
$\{x_1,x_2,\dots, x_n\}$ and $\mathcal{D}$ is a function mapping a variable $x_i\in\mathcal{X}$ to a domain of values $\mathcal{D}(x_i) = \{v_{i_1},v_{i_2}\dots
v_{i_{d_i}}\} $.  We note $d= max\{d_i| 1\leq i\leq n\}$ the maximum size of the domains, and ${\cal V}=\cup_{x\in\mathcal{X}} \mathcal{D}(x)$ the set of all values. $\mathcal{C}$ is a finite set of $m$ constraints
$\{c_1,c_2,\dots,c_m\}$. Each constraint $c_i \in \mathcal{C}$ of arity $k$ is
defined as a couple $\langle scope(c_i), R_{c_i}\rangle$ where $scope(c_i) = \{x_{i_1},
\ldots, x_{i_k}\}\subseteq\mathcal{X}$ is the set of variables involved in $c_i$
and $R_{c_i}\subseteq\mathcal{D}(x_{i_1}) \times \ldots \times \mathcal{D}(x_{i_k})$ the set of
allowed tuples i.e. $t\in R_{c_i}$ iff the tuple $t$ satisfies the constraint
$c_i$. We define the size of the constraint network $\mathcal{P}$ as $|\mathcal{P}|= \sum_{c\in\mathcal{C}} |R_{c}|$ where $|R_c| = \sum_{t\in R_c} |t|$ and $|t| = |scope(c)|$. A solution to the constraint network $\mathcal{P}$ is an assignment of all the variables satisfying all the constraints  in $\mathcal{C}$. A CSP (Constraint Satisfaction Problem) is the problem of deciding if a constraint network $\mathcal{P}$ admits a solution or not. 

We denote $t[x]$ the value of the variable $x$ in the tuple $t$. 
%A CSP $\mathcal{P}$ is called binary iff $\forall c_i\in\mathcal{C},
%|scope(c_i)|\leq 2$. For simplicity reason, a constraint $c$ with $scope(c) =
%\{x_i, x_j\}$ is denoted $c_{ij}$. \\
Let $t_1=(v_1,\ldots{},v_k)$ and $t_2=(w_1,\ldots{}, w_l)$ be two tuples (of values or variables), we define the non-commutative operator $\oplus$ by 
$t_1\oplus t_2 = (v_1,\ldots{}, v_k, w_1,\ldots{},w_l)$. Let $\mathcal{P} =\langle\mathcal{X}, \mathcal{D}, \mathcal{C}\rangle$ be a CSP instance, $c=\langle scope(c), R_c\rangle\in\mathcal{C}$ a constraint  and $s=(x_1,\ldots{},x_k)$ a sequence of variables such that  $Var(s)\subseteq scope(c)$ where $Var(s)$ is the set of variables of $s$.
We denote by $R_{c}{[s]}$ the following set of tuples: $$R_{c}{[s]}=\{(t[x_1],\ldots{},t[x_k])\mid t\in R_c \}$$

\subsection{Tseitin's Extension principle}
To explain the Tseitin principles \cite{Tseitin68} at the basis of linear transformation of general Boolean formulas to a formula in  conjunctive normal form (CNF), let us introduce some necessary definitions and notations. A {\em CNF formula}  $\Phi$  is a
conjunction of {clauses}, where a {\em clause} is a disjunction of {literals}. A {\em literal} is a positive ($p$) or negated ($\neg{p}$) 
propositional variable.  The two literals $p$ and $\neg{p}$ are called {\it complementary}. A CNF formula can also be seen as a set of clauses, and a clause as a set of literals. The size of the CNF formula $\Phi$ is defined as $|\Phi| = \sum_{c\in\Phi} |c|$, where $|c|$ is equal to the number of literals in $c$. 

%The size of the CNF formula $\Phi$ is defined as $|\Phi| = \sum_{c\in\Phi} |c|$, where $|c|$ is equal to the number of literals in $c$. 
%We denote by $\bar{l}$  the complementary literal of $l$. More precisely, if  $l = p$ then $\bar{l}$ is $\neg p$ and if  $l = \neg p$ then $\bar{l}$ is $p$.
%%For a set of literals $L$, $\bar{L}$ is defined as $\{\bar{l} ~|~ l \in L\}$. 
%% A CNF formula can also be seen as a set of clauses, and a clause as a set of literals. 
%Let us recall that any propositional formula can be translated to CNF using Tseitin's linear encoding \cite{Tseitin68}. We denote by ${\cal V}_{\Phi}$ the set of propositional variables appearing in $\Phi$, while the set of literals of $\Phi$ is defined as ${\cal L}_{\Phi}= \cup_{x\in {\cal V}_{\Phi} }\{x,\neg x\}$. An {\it interpretation} ${\cal B}$ of a propositional formula $\Phi$ is a function which  associates a value ${\cal B}(p)\in\{0, 1\}$ ($0$ corresponds to $false$ and $1$ to $true$)
%to the variables $p \in {\cal V}_{\Phi}$.  A {\it model} of a formula $\Phi$ is an  interpretation ${\cal B}$ that  satisfies the  formula:  ${\cal B}(\Phi)=1$. 
%The {\it SAT problem} consists in deciding if a given CNF formula admits a model or not. 
%

Tseitin's  encoding consists in introducing fresh variables to represent sub-formulae in order  to 
represent their truth values. Let us consider the following DNF formula (Disjunctive Normal Form: a disjunction 
of conjunctions): $$(x_1\lconj\cdots{}\lconj x_l)\ldisj (y_1\lconj\cdots{}\lconj y_m)\ldisj (z_1\lconj\cdots{}\lconj z_n)$$
A naive way of converting such a formula to a CNF formula consists in using the distributivity 
of disjunction over conjunction ($A\ldisj (B\lconj C)\leftrightarrow (A\ldisj B)\lconj (A\ldisj C)$):
$$(x_1\ldisj y_1\ldisj z_1)\lconj (x_1\ldisj  y_1 \ldisj z_2)\lconj\cdots{}\lconj (x_l\ldisj y_m\ldisj z_n) $$
Such a naive approach is clearly exponential in the worst case. In Tseitin's transformation, fresh propositional 
variables are introduced to prevent such combinatorial explosion, mainly caused by the distributivity of disjunction over conjunction and vice versa. With additional variables, the obtained CNF formula is linear in the size of the original formula. However the equivalence is only preserved w.r.t satisfiability:
$$(t_1\ldisj t_2\ldisj t_3)\lconj (t_1\rightarrow (x_1\lconj\cdots{}\lconj x_l)) \lconj (t_2\rightarrow (y_1\lconj\cdots{}\lconj y_m))$$
$\lconj (t_3\rightarrow (z_1\lconj\cdots{}\lconj z_n))$

\section{Compressing Table Constraints Networks}
In this section, we proposed two compression rules for table constraints networks. The first one is based on the constraint graph aims to reduce the size of the constraint network by rewriting the constraints using the most shared variables. The second compression technique based on the microstructure of the constraint network  aims to reduce the size of table constraints  by exploiting common sub-tuples.

\subsection{Constraint graph  Based Compression}
\subsubsection{CSP instance as transactions database:} We describe the transactions database that we associate to a given constraints network. It is obtained by considering the set of variables as a set of items.
\begin{definition}
Let ${\cal P}=\langle\mathcal{X}, \mathcal{D},\mathcal{C} \rangle$ be a constraints network.  The transactions  database associated to $\cal P$, denoted ${\cal TD}_{\cal P}$, 
is defined over the set of items $\cal X$ as follows: 
$${\cal TD}_{\cal P}=\{(tid_c, scope(c))\mid c\in C\} $$
\end{definition}

\subsubsection{Constraints Graph Rewriting Rule (CGR):}
%%%%%%%%%%%%%%%%%%%%%%%%
We provide a rewriting rule for reducing the size of a constraints network.  
It is mainly based on introducing new variables using Tseitin extension principle.

\begin{definition}[CGR rule]
\label{def:cgr}
Let $\cal P=\langle X, D, C\rangle$ be a constraints network, $s=(x_1,\ldots{}, x_k)$ a tuple of variables and 
$\{c_1, c_2,\ldots{}, c_n\}\subseteq \cal C$ a subset of $n$ constraints of $C$ such that ${\cal V}(s)\subseteq scope(c_i)$ for $1\leq i\leq n$. 
In order to rewrite $\cal P$,  we introduce a new variable $y\notin {\cal X}$ and a set $\mathcal{N}$ of $l$ new values such that $\mathcal{V}\cap\mathcal{N}=\emptyset$
and $l=|\bigcap_{i=1}^n R_{c_i}[s]|$. Let $f$ be a bijection from $\bigcap_{i=1}^n R_{c_i}[s]$ to $\mathcal{N}$.
We denote by $\mathcal{P}^g$ the constraint network $\langle \mathcal{X}^g, \mathcal{D}^g, \mathcal{C}^g\rangle $  obtained 
by rewriting  $\mathcal{P}$ with respect to $s$ and $f$: 
\begin{itemize}
\item $\mathcal{X}^g = \mathcal{X}\cup \{y\}$; 
\item $D^g$ is a domain function defined as follows: $\mathcal{D}^g(x) = \mathcal{D}(x)$ if $x\in\mathcal{X}$, and $\mathcal{D}^g(y) = \mathcal{N}$. 
\item $\mathcal{C}^g = \mathcal{C}\setminus \{c_1,\ldots{},c_n\}\cup C'$ , where $C'=\{c_0, c_1',\ldots{}, c_n'\}$ such that:
\begin{itemize}
\item $c_0 =\langle (y,x_1,\ldots{},x_k), \{(f(a_1,\ldots{},a_k), a_1,\ldots{},a_k)| (a_1,\ldots{},a_k)\in \bigcap_{i=1}^n R_{c_i}[s]\}\rangle$
%\item $c_i'=\langle (scope(c_i) -s)\oplus (y), \{\alpha\oplus (f(\beta)) | \alpha\oplus \beta \in R_{c_i}[(scope(c_i)-s)\oplus s]\}\rangle $
\item $c_i'=\langle (scope(c_i) -s)\oplus (y), \{t[scope(c_i)-s]\oplus (f(t[s])) |t \in R_{c_i}, t[s]\in\bigcap_{j=1}^n R_{c_j}[s]\}\rangle $
\end{itemize}
\end{itemize}
\end{definition}

It is important to note that our rewriting rule, achieve a weak form of pairwise consistency \cite{Janssen89}.  A constraint network is pairwise consistent (PWC) iff it has non-empty relations and any consistent tuple of a constraint $c$ can be consistently extended to any other constraint that intersects with $c$. 

\begin{definition}[Pairwise consistency]\cite{Beeri83,Janssen89}
Let $\cal P=\langle X, D, C\rangle$ be a constraints network. $\cal P$ is pairwise consistent if and only if $\forall c_i, \forall c_j\in\mathcal{C}, R_{c_i}[scope(c_i)\cap scope(c_j)] = R_{c_j}[scope(c_i)\cap scope(c_j)]$ and $\forall c\in{\cal C}$, $R_c\neq\emptyset$.
\end{definition}

As pairwise consistency deletes tuples from a constraint relation, some values can be eliminated when they have lost all their supports. Consequently, domains can be filtered if generalized arc consistency (GAC) is applied in a second step. 

As a side effect, our CGR rewriting rule maintains some weak form of PWC. Indeed, in Definition \ref{def:cgr},  when a sub-tuple $t[s]\notin \bigcap_{j=1}^n R_{c_j}[s]$, the  tuple $t$ is then deleted and do not belong to the new constraint $c'_i$.

\begin{example} 
\label{ex:red}
Let $\cal P=\langle X, D, C\rangle$ be a constraints network, 
where $\mathcal{X} = \{x_1,\dots, x_4\}$, $\mathcal{D}(x_1) =,\dots, = \mathcal{D}(x_4) = \{a, b\}$ 
and $\mathcal{C} = \{c_1, c_2\}$ where $c_1 = \langle\{x_1, x_2, x_3\}, \{(b, a, a),$\\$ (a, a, b), (a, b, a)\}\rangle$ and $c_1 = \langle\{x_2, x_3, x_4\}, \{(a, b, a), (b, a, a), (b, a, b)\}\rangle$. Let $s= (x_2, x_3)$ be a tuple of variables such that $s\subset scope(c_1)$ and $s\subset scope(c_2)$.By applying the CGR rule on $\mathcal{P}$, we obtain $\mathcal{P}^g = \langle \mathcal{X}^g, \mathcal{D}^g, \mathcal{C}^g\rangle $ such that:
\begin{itemize}
\item $\mathcal{X}^g = \mathcal{X}\cup\{y\}$
\item $\forall i (1\leq i\leq 4), \mathcal{D}^g(x_i) = \{a, b\}$. We have  $\bigcap_{j=1}^2 R_{c_j}[s] = \{(a, b), (b, a)\}$. We define $f((a, b)) = c, f((b, a)) = d$. Then $\mathcal{D}^g(y) = \{c, d\}$.
\item $\mathcal{C}^g = \{c_0, c'_1, c'_2\}$
\begin{itemize}
\item $c_0 = \langle\{y, x_2, x_3\}, \{(c, a, b), (d, b, a)\} \rangle$;
\item $c'_1 = \langle\{x_1, y\}, \{(a, c), (a, d )\} \rangle$ and $c'_2 = \langle\{x_4, y\}, \{(a, c), (a, d ), (b, d)\} \rangle$
\end{itemize}
\end{itemize}
In this simple example, using one additional variable, we reduce the size of the constraint network from $|\mathcal{P}| = 18$ to $|\mathcal{P}^g| = 16$. As we can observe, the value $b$ can be eliminated by GAC from the domain of $x_1$.
\end{example}

\subsubsection{Necessary and sufficient condition for size reduction}
\vspace{0.3cm}

Let $\cal P=\langle X, D, C\rangle$ be a constraints network, and $s = (x_{1}, \dots, x_{n})\subseteq\mathcal{X}$ be a sub-tuple of variables corresponding to a frequent itemset $I_s$ of $\mathcal{P}^g$ where the minimal support threshold is greater or equal to  $k$. Let $\{c_1, \dots, c_k\}\subseteq\mathcal{C}$ be the set of constraints such that $s\subseteq scope(c_i)$ for $1\leq i\leq k$.   Suppose that the constraints network $\cal P$ is pairwise consistent, in such a case, all the relations associated to each $c_i$ for $1\leq i\leq k$ contain the same number $p$ of tuples.  Under such worst case hypothesis, the size of $\cal P$ can be reduced by at least $r = (n\times p \times k - (p\times k + n\times p + p)$. Let us consider again the example \ref{ex:red}. The reduction is at least $r = (2\times 3\times 2) - (3\times 2 + 2\times 3 +3) = 12-15 = -3$. If we consider, the tuple $(b, a, b)\in R_{c_1}$ eliminated by the application of the CGR rule. This results in subtracting $5$ from the second term of $r$. Consequently, we obtain a reduction of $2$. 

Regarding the value of $k$, one can see that the compression is interesting when $r>0$ i.e.  $k> \frac{n+1}{n-1}$. Indeed, if $n< 2$ then there is no reduction. Thus, there are three cases : if $n=2$,  then $k\geq 4$, else if $n=3$ then $k\geq 3$, $k\geq 2$ otherwise. Therefore, the constraint network is always reduced when $k\geq 4$. We obtain exactly the same condition as in our mining based compression approach of Propositional CNF formula \cite{Mining4SATCore}. This is not surprising, as CGR rule is an extension of our Mining4SAT approach \cite{Mining4SATCore}  to CSP.

\subsubsection{Closed vs. Maximal:} In \cite{Mining4SATCore}, we introduced two condensed representations of the frequent itemsets: closed and maximal. We know that the set of maximal frequent itemsets is included in that of the closed ones. 
Thus, a small number of fresh variables and new clauses are introduced using the maximal frequent itemsets.
However, there are cases where the use of the closed frequent itemsets is more suitable. The example given in \cite{Mining4SATCore}, show the benefit that can be obtained by considering frequent closed itemsets. In our Mining for CSP approach we search for frequent closed itemsets.

\subsubsection{Compression algorithm:} Given a constraint network $\mathcal{P}$,  we first search for closed frequent itemsets (set of variables) on $\mathcal{TD}_{\mathcal{P}}$ and then we apply the above rewriting rule on the constraint network using the discovered itemsets  of variables. For more details on our algorithm, we refer the reader to the Mining4SAT greedy algorithm \cite{Mining4SATCore}, where the overlap notion between itemsets are considered.  The general compression problem can be stated as follows: given a set of frequent closed itemsets (sub-sequence of variables) and a constraints network, the question is to find an ordered sequence of operations (application of the CGR rule) leading to a CSP of minimal size. 

\subsection{Microstructure Based Compression}
In this section, we describe our compression based approach of Table constraints. First, we show how a Table constraint $c$ can be translated to a transaction database $\mathcal{TD}_c$. Secondly, we show how to compress $c$ using itemset mining techniques. 
\subsubsection{Table constraint as transactions database:}
Obviously, a table constraint $c$ can be translated in a naive way to a transaction database $\mathcal{TD}_c$.  Indeed, one can define the set of items as the union of the domains of the variables in the scope of $c$ ($\mathcal{I} = \cup_{x\in scope(c)}\mathcal{D}(x)$) and a transaction  $(tid, t)$ as the set of values involved in the tuple $t\in R_c$. This naive representation is difficult to exploit in our context. Let $I = \{a, b, c\}$ be a frequent itemset of $\mathcal{TD}_c$. As the variables in each transaction (or tuple) associated to the values in $I$ are different, it is difficult to compress the the constraint while using both classical tuples and compressed tuples \cite{KatsirelosW07}. To overcome this difficulty, we consider tuples as sequence, where each value is indexed by its position in the tuple. 

%\begin{definition}
%Let ${\cal P}=\langle\mathcal{X}, \mathcal{D},\mathcal{C} \rangle$ be a constraint network, and $c_i \in\mathcal{C}$ a table constraint such that $scope(c_i) = (x_{i_1}, x_{i_2},\dots, x_{i_{n_i}})$.  Let $t \in R_{c_i}$  a tuple of $c_i$. We define $indexed(t) = (t[x_{i_1}]^1, t[x_{i_2}]^2, \dots, t[x_{i_{n_i}}]^{n_i})$ as an indexed tuple associated to $t$ i.e. each value of the tuple is indexed with its position in the tuple. 
%\end{definition}

\begin{definition}[Indexed tuples]
Let ${\cal P}=\langle\mathcal{X}, \mathcal{D},\mathcal{C} \rangle$ be a constraint network, and $c_i \in\mathcal{C}$ a table constraint such that $scope(c_i) = (x_{i_1}, x_{i_2},\dots, x_{i_{n_i}})$.  Let $t \in R_{c_i}$  a tuple of $c_i$. We define $indexed(t) = (t[x_{i_1}]^1, t[x_{i_2}]^2, \dots, t[x_{i_{n_i}}]^{n_i})$ as an indexed tuple associated to $t$ i.e. each value of the tuple is indexed with its position in the tuple. 
\end{definition}

\begin{definition}[Inclusion, index]
Let $c$ be a table constraint with $scope(c) = \{x_1,\dots, x_n\}$ and $t =(v_1, \dots, v_n)\in R_c$ a tuple of $c$. We say that $s = (w_1,\dots, w_k)$ is a sub-tuple of $t$, denoted $s\subseteq t$,  if $\exists 1\leq i_1<i_2<\dots < i_k\leq n$ 
such that $w_1=v_{i_1}, \dots, w_k = v_{i_k}$.  We define $index(t) = \{1,\dots,n\}$, while $index(w) = \{i_1, \dots, i_k\}$. We also define $vars(index(t)) = scope(c)$ and $vars(index(w)) = \{x_{i_1}, \dots, x_{i_k}\}$.
\end{definition}

%\begin{definition}[Inclusion]
%Let $s = (w_1,\dots, w_k)$ and $t =(v_1\dots v_n)$ be two tuples. We say that $s$ is a sub-tuple of $t$, denoted $s\subseteq t$,  if $\{w_1,\dots,w_k\}\subseteq \{v_1,\dots,v_n\}$.  
%\end{definition}

\begin{definition}
Let ${\cal P}=\langle\mathcal{X}, \mathcal{D},\mathcal{C} \rangle$ be a constraints network, and $c\in\mathcal{C}$ a table constraint where $scope(c) = \{x_1,\dots, x_n\}$. The transaction  database associated to $c$, denoted ${\cal TD}_{c}$, 
is defined over the set of items $\mathcal{I} = \bigcup_{t\in R_c} \{t[x_1]^1, \dots, t[x_n]^n\}$ as follows: 
$${\cal TD}_{c}=\{(tid_{t}, indexed(t))| t\in R_c \} $$
\end{definition}

\begin{example}
 \label{ex:cdb}
Let ${\cal P}=\langle\mathcal{X}, \mathcal{D},\mathcal{C} \rangle$ be a constraints network, where $\mathcal{X} = \{x_1, x_2, x_3, x_4\}$, $\mathcal{D}(x) =  \mathcal{D}(y) = \mathcal{D}(z) = \mathcal{D}(t) = \{a,b\}$. Let $c\in\mathcal{C}$ a table constraint, such that $scope(c) = \{x_1, x_2, x_3, x_4\}$ and $R_c = \{(a, b, b, a), (a, a, b, b), (a, b, a, a),  (b, b, a, a),$\\$ (b, b, b, a)\}$. The transaction database $\mathcal{TD}_c$ associated to $c$ is defined as follows:

\begin{table}
\label{tab:td}
\begin{center}
\begin{tabular}{|c|c|}
\hline
tid & itemset \\
\hline
001 &  $a^1, b^2, b^3, a^4$ \\
\hline
002 & $a^1, a^2, b^3, b^4$\\	
\hline
003 &  $a^1, b^2, a^3, a^4$\\
\hline
004 & $b^1, b^2, a^3, a^4$ \\
\hline
005 & $b^1, b^2, b^3, a^4$ \\
\hline
\end{tabular}
\end{center}
\caption{$\mathcal{TD}_c$ a transaction database associated to $c$}
\end{table}
Let $I = \{b^2, a^4\}$ be an itemset of  $\mathcal{TD}_c$.  We have $\mathcal{S}(I, \mathcal{TD}_c) = |\{001, 003, 004, 005\}|=2$, $index(I) = \{2, 4\}$ and $vars(index(I)) = \{x_2, x_4\}$.  
\end{example}

\subsubsection{Microstructure Rewriting Rule (MRR):} We now provide a rewriting rule for reducing the size of a table constraint.  
\begin{definition}[MRR rule]
Let ${\cal P}=\langle\mathcal{X}, \mathcal{D},\mathcal{C} \rangle$ be a constraints network and $c\in\mathcal{C}$ be a table constraint with $scope(c) = \{x_1,x_2,\dots x_n\}$ and $|R_c| =m$. Let $I=\{v_1^{i_1},\ldots{}, v_{k}^{i_k}\}$ be an itemset of $\mathcal{TD}_c$ and $Y = vars(index(I)) = \{x_{i_1},\dots, x_{i_k}\}$.  In order to rewrite $c$ using $I$,  we introduce a new variable $z\notin {\cal X}$ and a set $\mathcal{N}$ of $l$ new values such that $\mathcal{V}\cap\mathcal{N}=\emptyset$ and $l=|\bigcup_{t\in R_c} t[Y]|$. Let $f$ be a bijection from $\bigcup_{t\in R_c} t[Y]$ to $\mathcal{N}$.
We denote by ${\cal P}^m$ the constraints network $\langle \mathcal{X}^m, \mathcal{D}^m, \mathcal{C}^m\rangle $  obtained 
by rewriting $c$ with respect to $I$ and $f$: 
\begin{itemize}
\item $\mathcal{X}^m = \mathcal{X}\cup \{z\}$; 
\item $D^m$ is a domain function defined as follows: $\mathcal{D}^m(x) = \mathcal{D}(x)$ if $x\in\mathcal{X}$, and $\mathcal{D}^m(z) = \mathcal{N}$. 
\item $\mathcal{C}^m = C\setminus \{c\}\cup C'$ , where $C'=\{c_0, c'\}$ such that:
\begin{itemize}
\item $c_0 =\langle (z,Y), \{(f(a_1,\ldots{},a_k), a_1,\ldots{},a_k)| (a_1,\ldots{},a_k)\in \bigcup_{t\in R_c} t[Y]\}\rangle$
%\item $c_i'=\langle (scope(c_i) -s)\oplus (y), \{\alpha\oplus (f(\beta)) | \alpha\oplus \beta \in R_{c_i}[(scope(c_i)-s)\oplus s]\}\rangle $
\item $c'=\langle (scope(c) -Y)\oplus (z), \{t[scope(c)-Y]\oplus (f(t[Y])) |t \in R_{c}\}\rangle $
\end{itemize}
\end{itemize}
\end{definition}

\begin{example} 
\label{ex:mrr}
Let us consider again the example \ref{ex:cdb}. Applying the $MR$ rewriting rule to $c$ with respect to $I =  \{b^2, a^4\}$, and $f$ where $f((b,a)) = c_1$ and   $f((a,b)) = c_2$, we obtain the following two constraints:  
\begin{itemize}
\item $c_0 = \langle\{z, x_2, x_4\}, \{(c_1, b, a), (c_2, a, b)\} \rangle$;
\item $c' = \langle\{x_1, x_3, z\}, \{(a, b, c_1 ), (a, b, c_2 ), (a, a, c_1), (b, a, c_1), (b, b, c_1) \} \rangle$
\end{itemize}
\end{example}
It is easy to see that in example \ref{ex:mrr}, applying MRR rule leads to a constraint of greater size. In what follows, we introduce a necessary and sufficient condition for reducing the size of the table constraint.

\subsubsection{Necessary and sufficient condition for size reduction}
\vspace{0.3cm}

Let $c$ be a table constraint, $p$ the number of tuples in $R_c$, and $s = (v_{1}, \dots, v_{n})\subseteq\mathcal{X}$ be a sub-tuple of values corresponding to a frequent itemset $I_s$ of $\mathcal{TD}^c$ where the minimal support threshold is greater or equal to  $k$. Let $\{t_1, \dots, t_k\}$ be the set of tuples such that $t_i[vars(index(s))] =s$ for $1\leq i\leq k$.  The size of $R_c$ can be reduced by at least $r = (n\times k - (p+ 1 +n + (p-k))$. Let us consider again the example \ref{ex:mrr}. The reduction is at least $r  = (2\times 4 - (5+ 1 + 2+ (5-4))= 8-9 = -1$. In this example, we increase the size of $c$ by one value. Indeed, $|R_{c}| = 20$ and $|R_{c_0}| + |R_{c'}| = 6 + 15 = 21$. 

Regarding the value of $k$, one can see that applying MRR rule is interesting when $r>0$ i.e.  $k> \frac{2\times p+n+1}{n+1}$. In the previous example, no reduction is obtained as $4> \frac{2\times 5+2+1}{2+1}$. ($4> 4$, the condition is not satisfied).   

\subsubsection{Compression algorithm of a table constraint:} Given a constraint network $\mathcal{P}$, and $c$ a constraint table of $\cal P$, we first search for closed frequent itemsets (sub-tuple of values) on $\mathcal{TD}_{c}$ and then we apply the above rewriting rule on the table constraint using the discovered itemsets  of values. Similarly to the constraint graph based compression algorithm, our microstructure based compression algorithm can be derived from the one defined in \cite{Mining4SATCore}.

As a summary, to compress general CSP, our approach first apply constraint graph based compression algorithm followed by the microstructure based compression algorithm.  
\section{Conclusion and Future Works}
In this paper, we propose a data-mining approach, called Mining4CSP, for 
reducing the size of constraints satisfaction problems when constraints are represented in extension.
It can be seen as a preprocessing step that aims to discover hidden structural knowledge
that are used to decrease the size of table constraints.
Mining4CSP combines both frequent itemset mining techniques for discovering interesting 
substructures, and Tseitin-based approach for a compact representation of Table constraints  
using these substructures. Our approach  is able to compact a CSP by considering both its associated constraint graph and microstructure. This allows us to define a two step algorithm. The first step, named coarse-grained compression, allows to compact the constraint graph using patterns representing subsets of variables. The second step, named fine-grained compression allows us to compact a given set of tuples of a given table constraint using patterns representing subset of values. 
Finally, an experimental evaluation on CSP instances is short term perspective. 

%Moreover, we will study the possibility of using other data-mining techniques for reduction in both size and solving time.

\bibliographystyle{plain}
\bibliography{satBib}
\end{document}